\documentclass[10pt,twocolumn,letterpaper]{article}

\usepackage[accsupp]{axessibility}
\usepackage[pagenumbers]{wacv} 

\usepackage{latexsym}
\usepackage{amsmath, amsthm, amssymb, amsfonts}
\usepackage{booktabs}
\usepackage{enumitem}
\usepackage{multirow}
\usepackage[dvipsnames]{xcolor}
\usepackage{pifont}
\usepackage{xr}
\usepackage{graphicx}
\usepackage{makecell}
\usepackage{comment}
\usepackage{color, colortbl}
\newcommand{\xmark}{\ding{55}}
\newcommand{\cmark}{\ding{51}}%
\newcommand{\lightcheck}{{\color{lightgray}\cmark}}

\usepackage{algorithm}
\usepackage{algpseudocode}

\usepackage[colorlinks = true,
            linkcolor = purple,
            urlcolor  = blue,
            citecolor = cyan,
            anchorcolor = black]{hyperref}

\usepackage{textcomp}

\usepackage[capitalize]{cleveref}
\crefname{section}{Sec.}{Secs.}
\Crefname{section}{Section}{Sections}
\Crefname{table}{Table}{Tables}
\crefname{table}{Tab.}{Tabs.}

\makeatletter
\begin{document}

\title{Training Strategies for Isolated Sign Language Recognition}

\author{
Karina Kvanchiani \\
{\tt\small karinakvanciani@gmail.com}
\and
Roman Kraynov \\
{\tt\small ranakraynov@gmail.com}
\and
Elizaveta Petrova \\
{\tt\small kleinsbotle@gmail.com}
\and
Petr Surovcev \\
{\tt\small petr.surovcev@gmail.com}
\and
Aleksandr Nagaev \\
{\tt\small sashanagaev1111@gmail.com}
\and
Alexander Kapitanov \\
{\tt\small kapitanovalexander@gmail.com}
\\
\\
\hspace{-25em}SberDevices, Russia
}

\maketitle

\def\slovoext{SlovoExt} 
\def\showingspeed{gesturing speed}
\def\cnns{CNNs}
\def\accuracy{top-1 accuracy} \def\speedup{speed up} 
\def\slowdown{slow down} 
\def\iouloss{IoU-balanced CE} 
\def\pretrained{pre-trained}

\begin{abstract}
\label{sec:abstr}
Accurate recognition and interpretation of sign language are crucial for enhancing communication accessibility for deaf and hard of hearing individuals. However, current approaches of Isolated Sign Language Recognition (ISLR) often face challenges such as low data quality and variability in gesturing speed. This paper introduces a comprehensive model training pipeline for ISLR designed to accommodate the distinctive characteristics and constraints of the Sign Language (SL) domain. The constructed pipeline incorporates carefully selected image and video augmentations to tackle the challenges of low data quality and varying sign speeds. Including an additional regression head combined with IoU-balanced classification loss enhances the model's awareness of the gesture and simplifies capturing temporal information. Extensive experiments demonstrate that the developed training pipeline easily adapts to different datasets and architectures. Additionally, the ablation study shows that each proposed component expands the potential to consider ISLR task specifics. The presented strategies enhance recognition performance across various ISLR benchmarks and achieve state-of-the-art results on the WLASL and Slovo datasets.
\end{abstract}

\section{Introduction}
\label{sec:intro}

Sign Languages are the primary means of communication for many deaf and hard of hearing individuals. According to data from the All-Russian Society of the Deaf (VOG\footnote{https://voginfo.ru/all-russian-society-of-the-deaf/}), there were more than 150,000 native speakers of Russian Sign Language (RSL) in 2019. Such languages do not replicate spoken language but possess their own lexicon and unique grammatical rules. Due to the language gap, deaf and hard of hearing people may experience prejudice in finding employment, pursuing academic education, or accessing medical services. Furthermore, learning Sign Language (SL) is challenging due to the limited number of teachers and native speakers.

However, developing robust Sign Language Recognition (SLR) systems remains challenging. Variations in gesturing speed, complex spatial-temporal features, and real-world capturing conditions (e.g., diverse backgrounds or lighting) can notably affect recognition accuracy. Furthermore, large-scale, high-quality datasets are still limited.

\begin{figure}[h]
\includegraphics[scale=0.123]{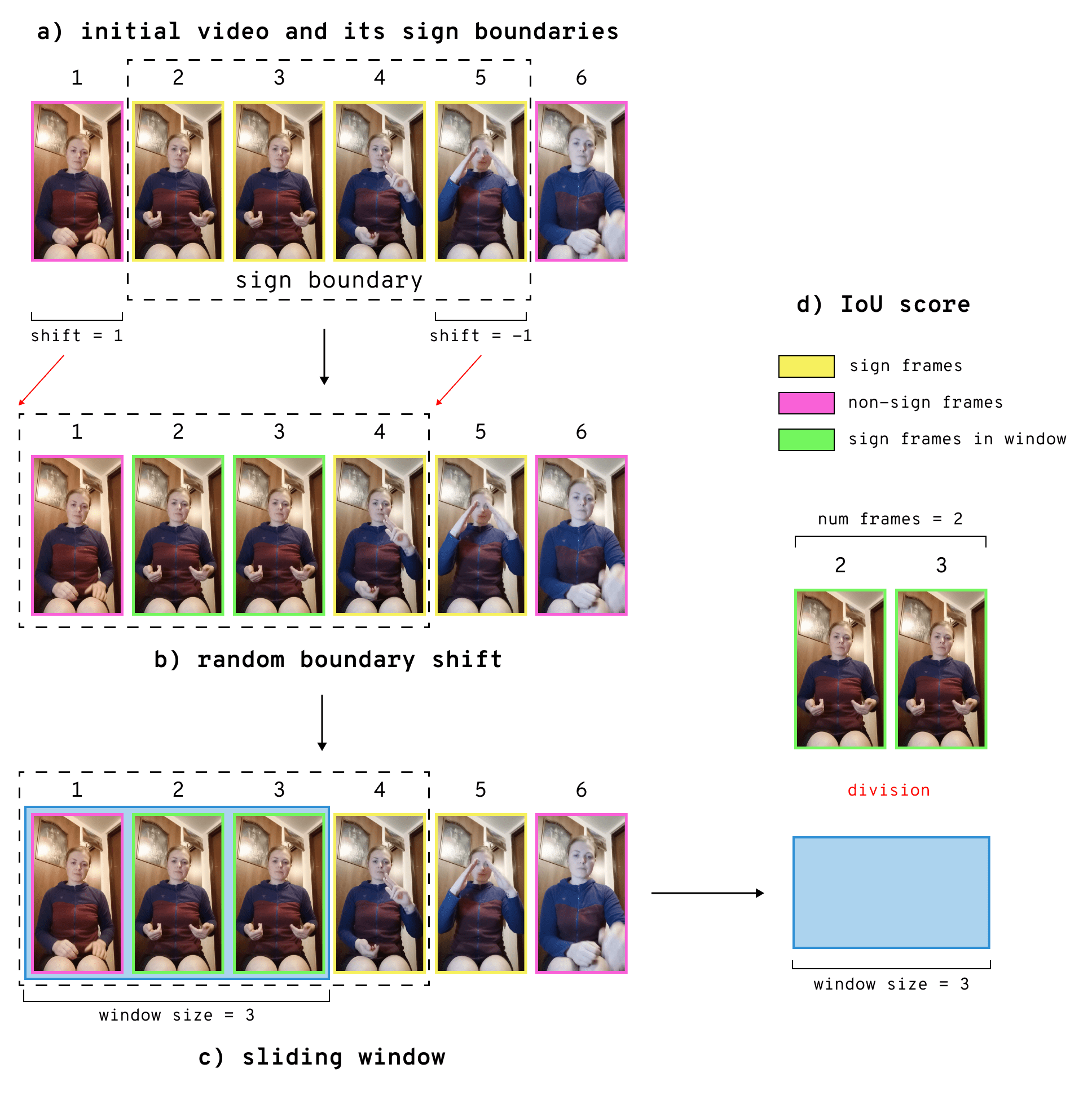}
\caption{The process of receiving IoU scores. (a) Consider the initial video, where the sign is between frames 2 and 5. (b) Randomly shift the sign`s boundaries by one frame to enhance model robustness (see Section~\ref{subsec:video_aug} for details). (c) Collect frames by sliding a window of size 3 across the video (window size of 3 chosen only for illustration). (d) Calculate IoU scores by dividing the number of sign frames in the window by the window size and adjust the classification scores. Note that the window size of 3 in this figure was selected purely as an example and was not used in our experiments.}
\label{fig:main}
\end{figure}

Sign language understanding is covered by three primary tasks: ISLR (Isolated Sign Language Recognition), CSLR (Continuous Sign Language Recognition), and SLT (Sign Language Translation). This research focuses on ISLR, where videos are classified into individual sign categories. One of the beneficial applications of this task is to create an automatic SL trainer, where users are shown a video example of a sign, and the recognition system assesses the quality of human gesturing. This advancement could make the learning of SL more accessible.

The development of automatic SLR systems is becoming widespread~\cite{slr_widespread, motivation} to facilitate communication between deaf and hearing individuals. SLR is a complex challenge due to the impact of hand movement, location, orientation, shape, and facial expressions on sign meaning~\cite{components}. The SLR system should function in real-world settings (schools, hospitals, or train stations). Its complexity is due to variations in sign display speeds, low video quality and resolution, diverse backgrounds, and varying lighting conditions. These factors adversely affect the system, which needs to operate with real-time response and maintain high quality to minimize errors. Besides real-world limitations, SLR faces domain-specific challenges such as limited data availability and complex temporal dependencies. Current methodologies often do not fully address these issues, limiting the development of models that generalize across diverse signing styles, varying execution speeds, and other conditions. To address these problems, we explore novel training strategies and propose a robust pipeline designed to handle the aforementioned challenges in the ISLR domain.

\begin{figure*}[h]
\includegraphics[scale=0.095]{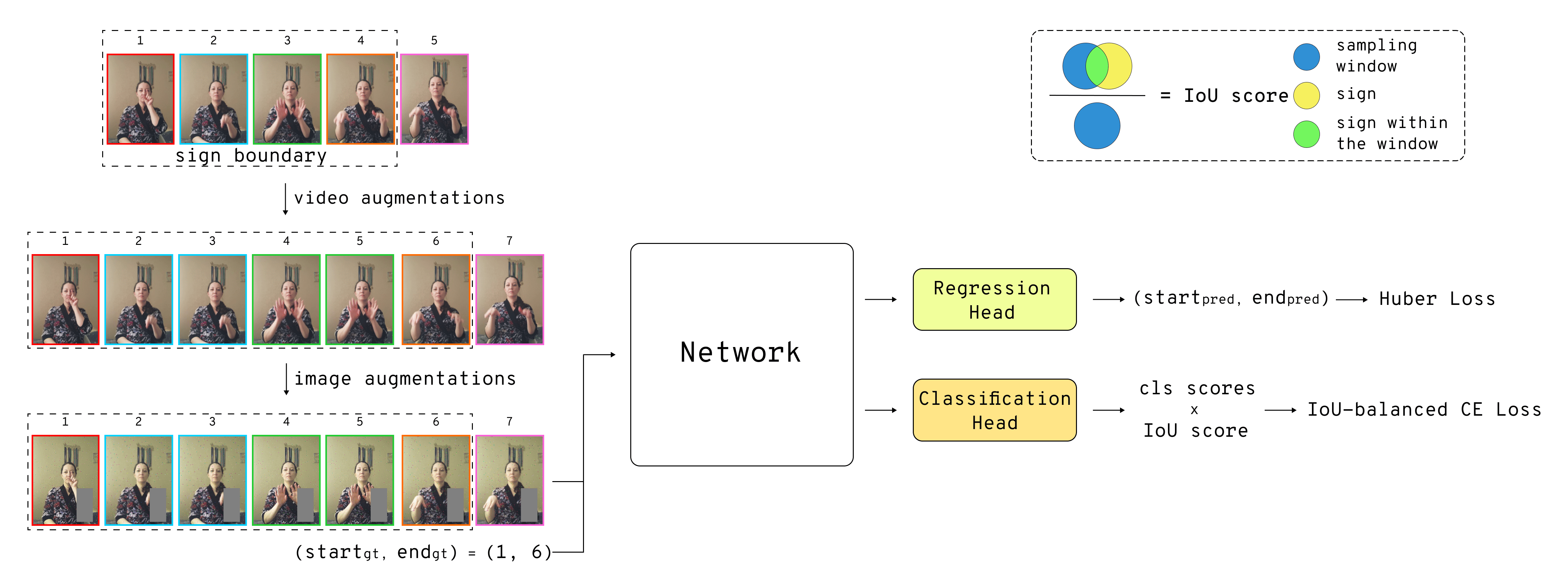}
\caption{Overall training pipeline. Video-level and image-level augmentations are applied, and the neural network is further trained with augmented data and sign boundary annotations.}
\label{fig:pipeline}
\end{figure*}

The basic approach to the ISLR task involves a classification neural network, designed to process RGB video data. The model takes the sequence of frames sampled from the video as input and predicts a text label corresponding to the sign depicted in the video. Many existing methodologies within this field provide either a solution at the model level~\cite{chen2023twostream, zhou2020spatialtemporal, hu2023continuous}, involving alterations to the model architecture, or at the data level~\cite{zuo2024improving}, focusing on expanding the training data. Due to unaddressed domain-specific challenges, these approaches may not fully leverage the models' potential for ISLR. Our work introduces a training pipeline designed for real-world SLR. It boosts performance without altering the underlying model architecture or dataset. To address both domain-specific challenges and practical deployment constraints, we apply a series of modifications to the basic approach. These enhancements improve accuracy across multiple datasets and model types, including transformers~\cite{mvitv2, maskfeat} and convolutional neural networks (\cnns)~\cite{i3d}: (1) video-level data augmentations to simulate the different gesturing speeds (shown in Figure~\ref{fig:video_aug}); (2) image-level data augmentations to reproduce low video quality; (3) an auxiliary sign boundary regression head to direct the network to focus more on the frames containing signs; and (4) 1D Intersection-over-Union-balanced CrossEntropy (\iouloss) loss to enhance the model's capability to understand signs (see Figure~\ref{fig:main}). Considering the necessity for real-time response, we focus on modifying strategies solely based on RGB data without additional modalities like hand keypoints~\cite{devisign} and depth~\cite{TheRuSLan, devisign} information. 

The contribution of this paper is fourfold:
\begin{itemize}
\item We propose a versatile and scalable training pipeline for ISLR models that considers domain-specific constraints, including data quality, varying sign execution speeds, and sign boundary awareness.
\item We present a novel large-scale dataset, \slovoext, which combines the Slovo~\cite{slovo} dataset with our newly assembled Russian Isolated Sign Language Dataset; we also release\footnote{\href{https://github.com/ai-forever/TrainingStrategiesISLR}{https://github.com/ai-forever/TrainingStrategiesISLR}} the code and \pretrained\ models to facilitate further research. The \slovoext\ dataset will be available as part of a larger RSL dataset in future work.
\item We illustrate how the proposed training strategies enable superior performance compared to existing solutions, bridging the gap between data-level and model-level improvements in SLR tasks.
\end{itemize}
\section{Related Work}
\label{sec:related_work}

\textbf{Sign Language Recognition Training Heuristics.} Many works in the SLR field have implemented model architecture-level alterations to improve performance. In \cite{self-mutual}, self-mutual distillation was employed to enable the model to learn temporal and spatial features. In \cite{hu2023continuous}, authors used correlation and identification modules to capture hands and face trajectories more effectively across consecutive frames. Other researchers have employed data-based approaches to boost their model. The authors of \cite{zuo2024improving} utilized datasets of various sign languages for training to address the issue of insufficient data. Some approaches have incorporated not only RGB but also pose information~\cite{camgoz2020multichannel}, hand and body keypoints~\cite{kan2021sign, chen2023twostream, zhou2020spatialtemporal}, and mouthing cues~\cite{prajwal2022weaklysupervised}, combining data- and model-level modifications. This paper focuses on developing a training pipeline to be applied to any ISLR model that takes RGB videos as input and outputs classification scores.

\textbf{Image \& Video Augmentations.}
Many studies focusing on SLR have employed random crop~\cite{chen2023simple, chen2023twostream, min2021visual, Ahn2023SlowFastNF, Pu_2020} and horizontal flip~\cite{self-mutual, min2021visual, Ahn2023SlowFastNF, Pu_2020} as standard image augmentations. However, these studies often do not address how to handle non-mirrored signs, which should be kept the same since altering them could change their meaning. Zhou et al. \cite{zhou2023glossfree} utilized strong data augmentation techniques, such as geometric and color space transformations, to enhance the model's robustness to data perturbations.  

Video augmentations have proven widely effective in the action recognition task to enhance the model's capacity to capture temporal dependencies. However, actions in this task (e.g., walking or running) are frequently repetitive and can be correctly classified at any time. On the contrary, the order of movements is essential in the SLR task. The specificity is the main reason for the inability to apply some of the transformation proposed in \cite{gorpincenko2022extending}, where authors augmented data with video reversing, frame mixing, and temporal extension of CutMix~\cite{yun2019cutmix}. In \cite{Pu_2020}, the authors proposed a video augmentation method where part of the video frames corresponding to one gesture is removed, added, or replaced with frames depicting a different gesture within the same video. This technique allowed the authors to address the data deficiency and leverage the contextual information within the sign sequence. In this paper, we only remove and add frames (``\speedup" and ``\slowdown" in Section~\ref{subsec:video_aug}) to speed adjustments because replacing is unsuitable for the ISLR task, given that only isolated signs are present. Ahn et al. \cite{Ahn2023SlowFastNF} used different frame sampling rates to capture spatial and temporal information separately. We adopt a similar approach with randomness (``random add" and ``random drop" in Section~\ref{subsec:video_aug}) to simulate real-life SLR cases. These video augmentations, shown in Figure~\ref{fig:video_aug}, enable sign recognition regardless of display speed and address data insufficiency in the SLR task.

\textbf{Auxiliary Regression Task.} Localizing the action boundaries via regression loss has demonstrated benefits in solving action recognition tasks. This approach enabled Zhang et al. \cite{zhang2022actionformer} to achieve single-stage anchor-free temporal action localization. Similarly, the authors of \cite{templocalization} divided the temporal action localization task into classification and action boundary regression, applying L1 loss for the latter. As was stated in \cite{liebel2018auxiliary}, supplementary tasks can boost the performance of the main task by pushing the backbone toward learning robust and generalized representations. Considering this concept, we incorporate an auxiliary sign boundary regression task by adding a regression head optimized by Huber loss~\cite{Huber1964RobustEO} to the training pipeline, as it has demonstrated better model convergence in the experiments provided below (see Table~\ref{tab:losses}). It assists the model in understanding the temporal position of the sign within the video, allowing it to focus more on the frames containing the sign.

\textbf{\iouloss\ Loss.} Extracting information about an object's spatial and temporal boundaries proves beneficial in both action recognition~\cite{XU2019351} and SLR tasks~\cite{jain2023addsl, prajwal2022weaklysupervised}. Techniques like IoU-loss~\cite{jain2023addsl, Liu_2022, wu2020ioubalanced} can aid this process. Wu et al. \cite{wu2020ioubalanced} proposed employing the \iouloss\ loss to address the independence between classification and localization predictions. Liu et al. \cite{Liu_2022} employed a 1D IoU loss in the action detection task to localize relevant segments within the video. In \cite{jain2023addsl}, the Generalized IoU loss was employed for SLR by calculating IoU between the bounding boxes of hands in each frame. In the proposed pipeline, we combine the concepts of 1D IoU and the IoU-balanced classification losses to devise the classification IoU loss tailored for the video domain and the particularities of the ISLR task. When computing the classification score, this approach allows us to consider the relative localization of the sign (see Figure~\ref{fig:main} for details).


\section{Training Strategies}
\label{sec:strategies}
To address the existing gap in SLR methodologies, designing an effective training pipeline requires tailored approaches that account for both data characteristics and model requirements. Figure~\ref{fig:pipeline} shows the proposed strategies divided into three components: (1) video-level augmentations, (2) image-level augmentations, and (3) additional losses. Note that these techniques do not depend on specific SL datasets or model architectures, except for certain subcomponents detailed in Section~\ref{subsec:image_aug}.

\subsection{Video Augmentation}
\label{subsec:video_aug}
Video augmentations are designed to reduce the gap between real cases and training data regarding subjects' \showingspeed by artificially changing the speed and shifting the sign boundaries.

\begin{figure}[h]
\includegraphics[scale=0.08]{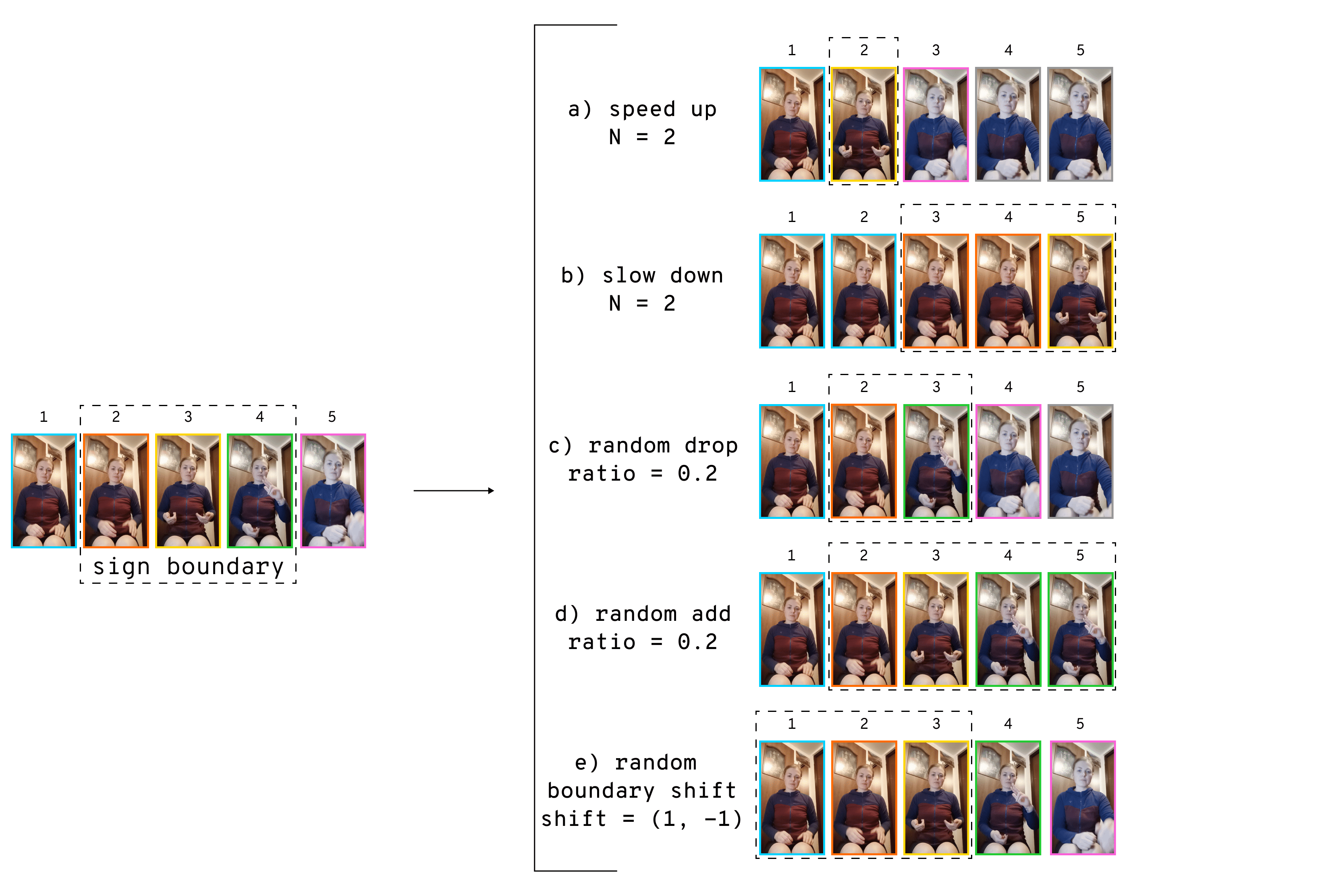}
\caption{Applying video augmentations to untrimmed videos. Identical frames are highlighted in the same color, and sign boundaries are outlined with a dashed line. Grey boundaries indicate duplicates of the last frame. a) \speedup\ the video 2 times by removing every second frame; b) \slowdown\ the video 2 times by duplicating every second frame; c) random frames drop remains 80\% of the total video length; d) random frames duplication increases the total video length by 20\%; e) random boundary shift is applied, e.g., with shift of (1, -1), which means one frame is added on the left and one is removed on the right. These values were chosen for ease of demonstration.}
\label{fig:video_aug}
\end{figure}

\textbf{Speed Up \& Slow Down.} In real life, signs can appear at various speeds, so we artificially \speedup\ or \slowdown\ videos to match the real-world \showingspeed\ distribution. We sample every N-th frame to \speedup\ by a factor of N (see Figure~\ref{fig:video_aug}a), pre-sampling additional frames on the right to maintain the same clip length. Conversely, we repeat each frame N times to \slowdown\ the video (see Figure~\ref{fig:video_aug}b), pre-dropping frames so that the final length remains unchanged.

\textbf{Random Add \& Random Drop.} In some real-life scenarios, \showingspeed\ is non-uniform because of the difficulty or simplicity of some gesture parts or due to external distractions during the hand movement. We partially change the speed of the videos to make the model more resistant to such uncommon cases. Figure~\ref{fig:video_aug}c illustrates the process of randomly dropping frames to \speedup\ the video partially (see Algorithm~\ref{alg:drop}). As in the case of uniform video \speedup, we retain the same clip length using the identical process. Similarly, a random add of frames is applied to the video for a partial \slowdown\ (see Figure~\ref{fig:video_aug}d). We are maintaining the video length in the same way as in a uniform \slowdown.

\textbf{Random Boundary Shift.} During inference, an SLR system generally captures frames sequentially using a window of a specific length. Therefore, signs that are too short or too long may not fit into the window properly. To ensure accurate recognition, we augment the data by randomly shifting sign boundaries, removing part of the videos, or adding frames without signs (see Figure~\ref{fig:main} and Figure~\ref{fig:video_aug}e for illustrations).
 
\begin{algorithm}
    \caption{Random Drop Realization}
    \label{alg:drop}
    \begin{algorithmic}[1]
        \State $y=[video_i]_{i=0}^{31}$\; \newline
        // Initialize the video sequence $y$ with frames indexed from 0 to 31
        \State $drop\_ratio \gets d; d \in (0, 1)$\; \newline
        // Set the drop ratio $d$, which determines the rate of frames to drop
        \State $y\_len \gets len(y)$\; \newline
        // Calculate the original length of the video sequence $y$
        
        \State $\hat{y}\_len \gets \frac{y\_len}{1 - drop\_ratio}$\; \newline
        // Compute the new length $\hat{y}\_len$ to extend the video sequence such that after dropping frames, the original length $y\_len$ is preserved

        \State $y.extend([video_k]_{k=32}^{\hat{y}\_len})$\; \newline
        // Extend the video sequence $y$ with additional frames indexed from 32 to $\hat{y}\_len$

        \State $\hat{y} \gets sort(random.choice(y, y\_len))$\; \newline
        // Randomly select $y\_len$ frames from the extended video sequence $y$ without repetitions. Sort the selected frames to maintain the original order of frames
    \end{algorithmic}
\end{algorithm}

\subsection{Image Augmentation}
\label{subsec:image_aug}
Due to the real-life limitations, the model must be adaptive to variations in background, subjects, image quality, and overall visual appearance. Image augmentations diversify the data in a frame-independent way to simulate real-life cases.

\textbf{Image Quality.} We imitate artifacts in video recordings caused by low-resolution capturing, due to defects in the video camera, or problems with video transmission over the network by compressing and downscaling images and adding random noise and sharpness. 

\textbf{Horizontal Flip.} Most meanings of the signs remain unchanged after horizontal mirroring. Hence, horizontal flip transformation facilitates the equal processing of signs shown by different hands. However, some signs are not invariant to the horizontal flip (e.g., RSL gestures such as ``left'', ``right'', ``heart'', and ``liver'' stop transferring their meanings if shown with the inappropriate hand). In the experiments, this augmentation only affects mirrored signs (more details in Section~\ref{subsec:preproc}). 

\textbf{General.} In addition to domain-specific augmentations, we utilize general ones to diversify the dataset: color jittering to provide heterogeneity over the color context, and CutMix~\cite{yun2019cutmix} and MixUp~\cite{zhang2018mixup} to induce the model to learn more generalizable features.

\subsection{Additional Losses}
\label{subsec:modified_technicues}
The following domain-specific training techniques incorporate domain knowledge of the sign boundaries into the model by scaling the classification loss and solving an additional task to enhance the main task's solution accuracy. These modifications are not model-level, as they can be integrated into either architecture.

\textbf{Sign Boundary Regression Head.} The average \showingspeed\ of various signs may differ, and even the same sign may take a different number of frames, affecting data distribution. Motivated by this, we add the regression head with a fully connected layer parallel to the classification head. This auxiliary head predicts the start and end of a sign in the input video to embed an implicit understanding of the length and \showingspeed\ into the model's backbone. To train the regression head, we utilize Huber loss~\cite{Huber1964RobustEO}, which combines the strengths of both MSE and MAE losses: it offers sensitivity and robustness, being less affected by the influence of outliers.

\textbf{Classification Loss Scaling.} Conventional classification losses, like cross-entropy, have the limitation to not consider information about the localization of the sign in the video, which can cause irrelevant gradients to optimize. We use a \iouloss\ loss to incorporate information about the location of the sign relative to the window. To calculate IoU scores, the length of the intersection of the sign with the sampled window is divided by the size of the window:

\begin{equation}
  IoU score = \frac{\min(w_{end}, s_{end}) - \max(w_{start}, s_{start})}{w_{end} - w_{start} + 1},
  \label{eq:iou_score}
\end{equation}
where $s_{start}$, $w_{start}$ are the first frames of the sign and the sampled window, and $s_{end}$, $w_{end}$ are the last ones, respectively; $min(a, b)$ and $max(a, b)$ functions denote the minimum and maximum of two values, respectively (for illustration, see Figure~\ref{fig:main} and Figure~\ref{fig:pipeline}). The overall classification loss is the combination of classification and IoU scores.


\section{Datasets}
\label{sec:datasets}

\begin{table}
\centering
\scalebox{0.58}{
\begin{tabular}{|l|c|c|c|c|c|}
\hline
Dataset & Classes & Videos & Signers & Resolution & Language\\
\hline \hline
LSE-Sign~\cite{lse} & 2,400 & 2,400 & 2 & FullHD & Spanish\\
LSA64~\cite{lsa64} & 64 & 3,200 & 10 & FullHD & Argentinian\\
MS-ASL~\cite{msasl} & 1,000 & 25,513 & 222 & varying & American\\
TheRuSLan~\cite{TheRuSLan} & 164 & 13 & 13 & FullHD & Russian\\
K-RSL~\cite{krsl} & 600 & 28,250 & 10 & FullHD & Kazakh-Russian\\
FluentSigners-50~\cite{fluentsigners} & 278 & 43,250 & 50 & varying & Kazakh-Russian\\
\hline \hline
WLASL2000~\cite{wlasl} & 2,000 & 21,083 & 119 & varying & American\\
AUTSL~\cite{autsl} & 226 & 38,336 & 43 & 512 \texttimes\ 512 & Turkish\\
Slovo~\cite{slovo} & 1,001 & 20,400 & 194 & HD / FullHD & Russian\\
\slovoext\ (ours) & 1,001 & 51,000 & 241 & HD / FullHD & Russian\\
\hline

\end{tabular}}
\caption{The main characteristics of the existing ISLR datasets.}
\centering
\label{tab:datasets}
\end{table}
We assess the training strategies described above on three large-scale ISLR datasets: WLASL~\cite{wlasl}, AUTSL~\cite{autsl}, and Slovo~\cite{slovo}. They are the most diverse open-source data in terms of dataset contributors, performing gestures (signers), and contexts (see Table~\ref{tab:datasets}). Evaluating on these datasets alone is sufficient to confirm the pipeline's effectiveness.

\begin{itemize}
\item \textbf{WLASL} dataset consists of 21,083 RGB-based videos trimmed by sign boundaries. Each video contains only one sign in American Sign Language (ASL), and each sign is performed by at least 3 different signers in various dialects\footnote{There is more than one way to show one word. Therefore, intraclass diversity occurs.}. The dataset was recorded in a studio with solid-colored backgrounds. The WLASL contains non-mirrored signs, but since such signs in its vocabulary are not established, we estimate the impact of flip augmentation for all samples in Section~\ref{subsec:abl_image}.

\item \textbf{AUTSL} was designed to simulate real-life context, i.e., different indoor and outdoor environments and diverse lighting conditions. It contains 38,336 trimmed video samples with 512 \texttimes\ 512 frame resolution and 20 different backgrounds. The AUTSL is divided into 226 Turkish Sign Language (TSL) signs performed by 43 signers. It comprises numerous similar signs, so models that can extract complex information from the input data must be used for training. The situation with non-mirrored signs is identical to the WLASL's, so applying the horizontal flip to videos is analyzed below.

\item \textbf{Slovo} is the largest, most diversified, and the only publicly available RSL dataset. The dataset contains 20,400 HD and FHD untrimmed videos performed by 194 signers. It is divided into 1,001 classes, including the additional ``no event" class, which indicates videos where the signer is not performing a sign. Slovo was crowdsourced, featuring real-life conditions and non-invariant signs to horizontal flip.
\end{itemize}

\subsection{\slovoext\ Dataset}
We expand the Slovo dataset with a self-assembled 30,600 videos. The combination is called \slovoext\ and comprises 51,000 samples divided into the same 1,001 classes as Slovo (see Table~\ref{tab:datasets}). The process of \slovoext\ creation is constructed similarly to maintain the Slovo distribution characteristics such as video length, signer's appearance, and context heterogeneity identical. \slovoext\ is also utilized to evaluate the proposed training strategies.

The dataset was created by native speakers of RSL and interpreters proficient in RSL. We involved diverse contributors in the data collection process to address concerns about differences in sign presentation between deaf and hard of hearing people. This approach aims to help neural networks manage gesture variability and mitigate the effects of native signer bias~\cite{desai2024slbias}. 

In collecting the dataset, we carefully considered the specific features of RSL, ensuring that all data was recorded in a single dialect. The dataset was compiled using pre-recorded video templates provided by the All-Russian Society of the Deaf. An exam consisting of 20 questions assessed RSL proficiency for participation in dataset recording. Experts who scored at least 90\% were allowed to take on the tasks.

\begin{table}[tb]
\centering
\scalebox{0.68}{
\begin{tabular}{|c|c|c|cc|}
\hline
\multirow{2}{*}{Dataset} & \multirow{2}{*}{Model} & \multirow{2}{*}{Pretrain Task} & \multicolumn{2}{c|}{\accuracy} \\ 
\cline{4-5}
 &  &  & \multicolumn{1}{c|}{basic approach} & proposed pipeline \\ 
 \hline\hline
\multirow{4}{*}{WLASL} & \multirow{2}{*}{MViTv2-S} & MaskFeat & \multicolumn{1}{c|}{49.83} & \(56.37_{\color{Green}+6.54}\) \\ \cline{3-3} 
 &  & \multirow{3}{*}{Classification} & \multicolumn{1}{c|}{51.88} & \(57.17_{\color{Green}+5.29}\) \\ \cline{2-2} 
 & MViTv2-B &  & \multicolumn{1}{c|}{54.31} & \(57.33_{\color{Green}+3.02}\) \\ \cline{2-2}  
 & I3D &  & \multicolumn{1}{c|}{35.55} & \(36.38_{\color{Green}+0.83}\) \\ 
 \hline
\multirow{4}{*}{AUTSL} & \multirow{2}{*}{MViTv2-S} & MaskFeat & \multicolumn{1}{c|}{91.69} & \(95.62_{\color{Green}+3.93}\) \\ \cline{3-3} 
 &  & \multirow{3}{*}{Classification} & \multicolumn{1}{c|}{90.27} & \(95.05_{\color{Green}+4.78}\) \\ \cline{2-2} 
 & MViTv2-B &  & \multicolumn{1}{c|}{93.00} & \(95.75_{\color{Green}+2.75}\) \\ \cline{2-2} 
 & I3D &  & \multicolumn{1}{c|}{85.22} & \(87.81_{\color{Green}+2.59}\) \\ \hline
\multirow{4}{*}{Slovo} & \multirow{2}{*}{MViTv2-S} & MaskFeat & \multicolumn{1}{c|}{71.45} & \(81.57_{\color{Green}+10.12}\) \\ \cline{3-3} 
 &  & \multirow{3}{*}{Classification} & \multicolumn{1}{c|}{77.54} & \(80.97_{\color{Green}+3.43}\) \\ \cline{2-2} 
 & MViTv2-B &  & \multicolumn{1}{c|}{79.31} & \(81.34_{\color{Green}+2.03}\) \\ \cline{2-2} 
 & I3D &  & \multicolumn{1}{c|}{62.79} & \(63.82_{\color{Green}+1.03}\) \\ \hline
\multirow{4}{*}{\slovoext} & \multirow{2}{*}{MViTv2-S} & MaskFeat & \multicolumn{1}{c|}{81.55} & \(87.31_{\color{Green}+5.76}\) \\ \cline{3-3} 
 &  & \multirow{3}{*}{Classification} & \multicolumn{1}{c|}{83.36} & \(85.90_{\color{Green}+2.99}\) \\ \cline{2-2} 
 & MViTv2-B &  & \multicolumn{1}{c|}{84.14} & \(86.72_{\color{Green}+2.58}\) \\ \cline{2-2} 
 & I3D &  & \multicolumn{1}{c|}{77.30} & \(79.74_{\color{Green}+2.44}\) \\ \hline
\end{tabular}}
\caption{Evaluation results. We present two \accuracy\ metrics for each setup: the metric on the basic approach and the metric obtained via the proposed training pipeline. Two \pretrained\ models on the K400 dataset~\cite{kinetics} are utilized: MViTv2 trained on the classification task and MaskFeat trained in a self-supervised manner to reconstruct masked pixels. Green values show gains over the basic approach.}
\label{tab:exps}
\end{table}

\section{Experiments}
\label{sec:experiments}
The effectiveness of the training pipeline is assessed by comparing two metrics: utilizing (1) the basic approach\footnote{The basic approach is the same as the proposed pipeline, but without video and image augmentations, regression head, and classification loss balancing.} and (2) proposed strategies. We fine-tune three architectures: two transformers (MViTv2-S with different pre-training and MViTv2-B~\cite{mvitv2}) and one CNN (I3D~\cite{i3d} with a ResNet-50) -- on four ISLR datasets. The experiment results are provided in Table~\ref{tab:exps}.

The \accuracy\ is the primary metric for evaluation, measuring the percentage of videos where the predicted class matches the correct class. This metric was chosen due to its widespread use in the SLR task, facilitating comparison with other models and benchmarking. Mean accuracy was used to demonstrate state-of-the-art result on the Slovo dataset and ensure consistency with the metric on the leaderboard\footnote{https://paperswithcode.com/sota/sign-language-recognition-on-slovo-russian}.

\subsection{Preprocessing}
\label{subsec:preproc}
Input videos are resized to 300 resolution and converted to HDF5 video format. During the training stage, videos are square-padded and randomly cropped to achieve a resolution of 224 \texttimes\ 224. The low side of the videos is resized to 224 and square-padded during the validation and testing stages. The model input is a sequence of 32 frames sampled with a step of 2. If the video does not have enough frames for sampling, the last frame is repeated to form a complete clip. For the Slovo and \slovoext\ datasets, non-mirrored signs are excluded from the horizontal flip augmentation. The WLASL signs are not being flipped, and the AUTSL signs are flipped with no exceptions because Section~\ref{sec:ablation} shows the efficacy of such decisions. 

\begin{table*}[tb]
  \scalebox{0.66}{
  \begin{tabular}{|c|cc|ccccc|cc|c|}
  \hline
 \multirow{3}{*}{Ablation on} & \multicolumn{2}{|c|}{Image Augmentations} & \multicolumn{5}{|c|}{Video Augmentations} & \multicolumn{2}{|c|}{Additional Losses} & \\
  \cline{2-10}
   & \multirow{2}{*}{Basic} & \multirow{2}{*}{CutMix \& MixUp} & \multirow{2}{*}{Bound. Shift} & \multirow{2}{*}{Rand. Add} & \multirow{2}{*}{Speed Up} & \multirow{2}{*}{Slow Down} & \multirow{2}{*}{Rand. Drop} & \multirow{2}{*}{Huber Loss} & \multirow{2}{*}{\iouloss} & \multirow{1}{*}{top-1 acc.} \\
    & & & & & & & & & & \\
    \hline\hline
    None: entire pipeline & \cmark & \cmark & \cmark & \cmark & \cmark & \cmark & \cmark & \cmark & \cmark & \textbf{87.31}  \\
    \hline\hline
    \multirow{3}{*}{Image Augs.} & \xmark & \lightcheck & \lightcheck & \lightcheck & \lightcheck & \lightcheck & \lightcheck & \lightcheck & \lightcheck & \(86.08_{\color{red}-1.23}\) \\
     & \xmark & \xmark & \lightcheck & \lightcheck & \lightcheck & \lightcheck & \lightcheck & \lightcheck & \lightcheck & \(83.25_{\color{red}-4.06}\) \\
     & \lightcheck & \xmark & \lightcheck & \lightcheck & \lightcheck & \lightcheck & \lightcheck & \lightcheck & \lightcheck & \(85.42_{\color{red}-1.89}\) \\
    \hline\hline
    \multirow{7}{*}{Video Augs.} & \lightcheck & \lightcheck & \xmark & \xmark & \xmark & \xmark & \xmark & \lightcheck & \lightcheck & \(86.77_{\color{red}-0.54}\) \\
    & \lightcheck & \lightcheck & \xmark & \lightcheck & \lightcheck &\lightcheck & \lightcheck & \lightcheck & \lightcheck & \(87.00_{\color{red}-0.31}\) \\
    & \lightcheck & \lightcheck & \lightcheck & \xmark & \lightcheck & \lightcheck & \lightcheck & \lightcheck & \lightcheck & \(86.96_{\color{red}-0.35}\) \\
    & \lightcheck & \lightcheck & \lightcheck & \lightcheck & \xmark & \lightcheck & \lightcheck & \lightcheck & \lightcheck & \(86.99_{\color{red}-0.32}\) \\
    & \lightcheck & \lightcheck & \lightcheck & \lightcheck & \lightcheck & \xmark & \lightcheck & \lightcheck & \lightcheck & \(86.65_{\color{red}-0.66}\) \\
    & \lightcheck & \lightcheck & \lightcheck & \lightcheck & \lightcheck & \lightcheck & \xmark & \lightcheck & \lightcheck & \(87.01_{\color{red}-0.30}\) \\
    \hline\hline
    \multirow{4}{*}{Losses} & \lightcheck & \lightcheck & \lightcheck & \lightcheck & \lightcheck & \lightcheck & \lightcheck & \xmark & \lightcheck & \(86.79_{\color{red}-0.52}\) \\
     & \lightcheck & \lightcheck & \lightcheck & \lightcheck & \lightcheck & \lightcheck & \lightcheck & \xmark & \xmark & \(86.95_{\color{red}-0.36}\) \\
    & \lightcheck & \lightcheck & \lightcheck & \lightcheck & \lightcheck & \lightcheck & \lightcheck & \lightcheck & \xmark & \(87.26_{\color{red}-0.05}\) \\
    \hline
  \end{tabular}}
    \caption{Ablation study results. We employ the MViTv2-S model to assess the influence of each component by removing it from the pipeline and comparing the results. Accuracy is evaluated on the \slovoext\ combination. Red values indicate differences with the entire pipeline.}
\label{tab:ablation}
\end{table*}

\begin{table}[tb]
\centering
  \scalebox{0.9}{
  \begin{tabular}{|c|c|}
  \hline
    Method & \accuracy\ \\
    \hline\hline
     RandAugment & \(86.79_{\color{red}-0.52}\) \\
    \hline
    UniformAugment & \(86.75_{\color{red}-0.56}\) \\
    \hline
    Sequential (utilized) & 87.31  \\
    \hline
  \end{tabular}}
    \caption{Ablation study for different image augmentation methods.}
\label{tab:aug_types}
\end{table}

\begin{table}[tb]
\centering
  \scalebox{1}{
  \begin{tabular}{|c|c|}
  \hline
    Regression Loss & \accuracy\ \\
    \hline\hline
     MAE & \(86.68_{\color{red}-0.63}\) \\
    \hline
    MSE & \(87.00_{\color{red}-0.31}\) \\
    \hline
    Huber (utilized) & 87.31 \\
    \hline
  \end{tabular}}
    \caption{Ablation study for different regression losses.}
\label{tab:losses}
\end{table}

\subsection{Training Methodology}
\label{subsec:details}
As shown in Figure~\ref{fig:pipeline}, the preprocessed video is initially exposed to video augmentations. One randomly selected from four modifications is applied to each batch. In contrast, ``random boundary shift" is constantly affected. In experiments, ``\speedup" increases the video speed by 2 times, and ``\slowdown" decreases it by 2 times. ``Random drop" shortens video by 10\%, while ``random add" lengthen it by 30\%. The random values for sign boundaries shifting are from the interval [-5, 5], except the interval [-5, 0] for the WLASL and AUTSL datasets due to their trimming by the sign boundaries.

Image augmentations modify the video batch by randomly choosing a combination of them with expertly selected probabilities and magnitudes. Note that CutMix~\cite{yun2019cutmix} and MixUp~\cite{zhang2018mixup} influence only transformer models in our experiments because they can degrade \cnns\ due to a lack of correlation between neighboring pixels~\cite{cnn_cumix}.

We incorporate IoU scores into the classification head, multiplying classification scores by them. The additional regression head is fed with video features and sign boundaries. The network is trained by optimization \iouloss\ and Huber losses for classification and regression heads, respectively. Both are not applied to ``no event" videos in the Slovo and the \slovoext\ datasets by assigning an IoU score of one and sign boundaries of zeros. Although sign boundaries trim the WLASL and AUTSL datasets, the regression head works appropriately with them.

Training is performed until convergence on four Tesla H100s with 80GB RAM. Early stopping is triggered if the \accuracy\ metric does not increase by at least 0.003 after 7 epochs. For the first 20 epochs, the learning rate is modified by a linear scheduler, and a cosine scheduler is used from epochs 20 to 100. Cross-entropy loss is minimized using the AdamW~\cite{adamW} optimizer. Other parameters are variable and can be adjusted as needed.

\subsection{Results}
\label{subsec:results}

As Table~\ref{tab:exps} shows, the proposed pipeline consistently improves \accuracy\ across different datasets and architectures. Specifically, MViTv2-S achieves additional gains of 6.54\%, 3.93\%, 10.12\%, and 5.76\% on WLASL, AUTSL, Slovo, and \slovoext, respectively, while I3D obtains an average improvement of 1.72\% across these datasets. The ablation study (see Table~\ref{tab:ablation}) further underscores the effectiveness of each proposed component in the pipeline.

Moreover, this approach attains state-of-the-art performance on two benchmark ISLR datasets. On WLASL, using MViTv2-S \pretrained\ with MaskFeat and further pretrained on \slovoext, the model achieves 62.89\% \accuracy, surpassing the NLA-SLR~\cite{nlaslr} model by 1.63\%. On the Slovo dataset, this method reaches 78.21\% mean accuracy, exceeding the prior best result by 14.12\%. A comparable increase of 13.06\% is achieved with the MViTv2-S \pretrained\ on K400, confirming the robustness of the proposed training strategies.


\section{Ablation Study}
\label{sec:ablation}

This section estimates the impact of each part of the proposed pipeline individually. We divide all modifications into three blocks: (1) image augmentations, (2) video augmentations, and (3) additional losses. We test the necessity for each block by disconnecting it from the entire pipeline and for its parts by shutting down a particular part while not changing anything else. The ablation study uses MViTv2-S \pretrained\ on the MaskFeat to train on the \slovoext\ combination.

\subsection{Image Augmentations Necessity}
\label{subsec:abl_image}
The image augmentations block is additionally divided into two parts -- basic image transformations and CutMix-MixUp pair -- to simplify assessing their influence. Each part and combinations of them are shut down, resulting in three experiments. The first three rows in Table~\ref{tab:ablation} illustrate that the metric significantly decreases by 4.06\% in the absence of all image augmentations. The influence of basic image augmentation is two times less than the CutMix-MixUp one.

We also evaluate various augmentation-applying approaches, including RandAugment~\cite{randaugment}, UniformAugment~\cite{uniformaugment}, and random sequential applications. Table~\ref{tab:aug_types} indicates that RandAugment and UniformAugment approaches are not effective in the proposed pipeline. 

Additionally, since we are uncertain which signs are non-mirrored in WLASL and AUTSL datasets, we search for the optimal option via two experiments for each: one with flip augmentation and one without for all signs. On WLASL, the metric with a flip is lower by 1.3\%, indicating that this dataset probably contains numerous non-mirrored signs. On AUTSL, conversely, the metric with flip was higher by 0.11\%. Therefore, in the main experiments, we apply flip augmentation to the AUTSL dataset and refrain from using it with the WLASL dataset.

\subsection{Video Augmentations Necessity} 
\label{subsec:abl_video}
Similar to the above subsection, we split the estimation of video augmentations necessity by disconnecting each of them separately and the whole block, providing six experiments. The most notable impact achievable with one modification is produced by the ``\slowdown" augmentation, as without it, the \accuracy\ drops by 0.66\%. 

\subsection{Additional Losses Necessity} 
\label{subsec:abl_losses}
The process of evaluation is constructed identically. The last three rows in Table~\ref{tab:ablation} show that the regression loss provides a more substantial improvement than the \iouloss\ loss, resulting in a 0.52\% increase in metrics compared to 0.05\% increase. The result suggests that information about absolute sign boundaries is more crucial than relative sign position within the sampling window.

Also, we conduct additional research on the impact of Huber regression loss by replacing it with MAE and MSE losses (see Table~\ref{tab:losses}). We observe lower metrics in both cases, with MAE yielding the lowest at 86.68\%. These results could be attributed to the behavior of MAE, which, while descending rapidly, may get stuck on values close to the ground truth, overshooting the desired target value.


\section{Discussion}
\textbf{Ethical Considerations.} All participants signed consent forms before data mining. The forms authorized the processing and publication of personal data for research purposes. We do not restrict videos with signers under 18 since parental permission was obtained during the registration, which complies with the Civil Code of the Russian Federation\footnote{https://ihl-databases.icrc.org/en/national-practice/federal-law-no-152-fz-personal-data-2006}. To preserve contributors' privacy, we employ anonymized user hash IDs in the dataset annotations. Furthermore, we have ensured that the Slovo dataset meets these ethical criteria. We provide the dataset for research purposes only, but we understand that it could be misused for malicious purposes, such as identifying people or large-scale surveillance. 

\textbf{Positional Statement.} Through the research of the ISLR task, we involved the All-Russian Community of Deaf experts and professional sign language interpreters. The expertise of the All-Russian Society of the Deaf was utilized at every stage of the \slovoext\ dataset creation, including data collection, validation, and verification processes, as well as video tagging. We also involved deaf consultants in developing training strategies to apply Considerations to particular solutions. Additionally, some of our researchers took formal courses on RSL to enhance their knowledge of this domain. 

\textbf{Limitations.} The regression head and \iouloss\ loss require sign boundary annotations that are frequently difficult to achieve. Thus, when the dataset is pre-trimmed, the ``random boundary shift" augmentation can adjust the sign boundaries only inward toward the center rather than allowing shifts in both directions. We processed RGB frames and did not analyze articulation, keypoints, or depth information. Moreover, a lack of awareness regarding non-mirrored signs in sign language may lead to issues when applying flip augmentation. There are additional challenges in each sign language, and each has its specifics that must be considered when adapting the pipeline.

\section{Conclusion}
\label{subsec:conclusion}
In this paper, we introduce a training pipeline for the ISLR models, considering the specifics of the SLR domain and the constraints of real-world usage. We demonstrate the effectiveness of applying image and video augmentations to address the issues of low data quality and varying \showingspeed. The importance of integrating temporal information into the model by the regression head combined with \iouloss\ loss is also presented. The developed pipeline delivers state-of-the-art results on the WLASL and Slovo datasets. Future work will extend the training strategies to CSLR and SLT tasks.

\section*{Acknowledgements}
\label{sec:ack}
We are grateful to Alena Fenogenova, Albina Akhmetgareeva and Anastasia Vasyatkina for the discussions and comments on this work.
{\small
\bibliographystyle{ieee_fullname}
\bibliography{egbib}

\begin{thebibliography}{10}\itemsep=-1pt

\bibitem{Ahn2023SlowFastNF}
Junseok Ahn, Youngjoon Jang, and Joon~Son Chung.
\newblock \href{https://ieeexplore.ieee.org/abstract/document/10445841/}{Slowfast Network for Continuous Sign Language Recognition}.
\newblock In {\em ICASSP 2024-2024 IEEE International Conference on Acoustics, Speech and Signal Processing (ICASSP)}, pages 3920--3924. IEEE, 2024.

\bibitem{cnn_cumix}
Sihun Baek, Jihong Park, Praneeth Vepakomma, Ramesh Raskar, Mehdi Bennis, and Seong-Lyun Kim.
\newblock \href{https://federated-learning.org/fl-ijcai-2022/Papers/FL-IJCAI-22_paper_27.pdf}{Visual transformer meets cutmix for improved accuracy, communication efficiency, and data privacy in split learning}.
\newblock In {\em arXiv preprint arXiv:2207.00234}, 2022.

\bibitem{camgoz2020multichannel}
Necati~Cihan Camgoz, Oscar Koller, Simon Hadfield, and Richard Bowden.
\newblock \href{https://link.springer.com/chapter/10.1007/978-3-030-66823-5_18}{Multi-channel transformers for multi-articulatory sign language translation}.
\newblock In {\em Computer Vision--ECCV 2020 Workshops: Glasgow, UK, August 23--28, 2020, Proceedings, Part IV 16}, pages 301--319. Springer, 2020.

\bibitem{i3d}
Joao Carreira and Andrew Zisserman.
\newblock \href{http://openaccess.thecvf.com/content_cvpr_2017/html/Carreira_Quo_Vadis_Action_CVPR_2017_paper.html}{Quo vadis, action recognition? a new model and the kinetics dataset}.
\newblock In {\em proceedings of the IEEE Conference on Computer Vision and Pattern Recognition}, pages 6299--6308, 2017.

\bibitem{chen2023simple}
Yutong Chen, Fangyun Wei, Xiao Sun, Zhirong Wu, and Stephen Lin.
\newblock \href{http://openaccess.thecvf.com/content/CVPR2022/html/Chen_A_Simple_Multi-Modality_Transfer_Learning_Baseline_for_Sign_Language_Translation_CVPR_2022_paper.html}{A simple multi-modality transfer learning baseline for sign language translation}.
\newblock In {\em Proceedings of the IEEE/CVF conference on computer vision and pattern recognition}, pages 5120--5130, 2022.

\bibitem{chen2023twostream}
Yutong Chen, Ronglai Zuo, Fangyun Wei, Yu Wu, Shujie Liu, and Brian Mak.
\newblock \href{https://proceedings.neurips.cc/paper_files/paper/2022/hash/6cd3ac24cdb789beeaa9f7145670fcae-Abstract-Conference.html}{Two-stream network for sign language recognition and translation}.
\newblock In {\em Advances in Neural Information Processing Systems}, volume~35, pages 17043--17056, 2022.

\bibitem{randaugment}
Ekin~D Cubuk, Barret Zoph, Jonathon Shlens, and Quoc~V Le.
\newblock \href{http://openaccess.thecvf.com/content_CVPRW_2020/html/w40/Cubuk_Randaugment_Practical_Automated_Data_Augmentation_With_a_Reduced_Search_Space_CVPRW_2020_paper.html}{Randaugment: Practical automated data augmentation with a reduced search space}.
\newblock In {\em Proceedings of the IEEE/CVF conference on computer vision and pattern recognition workshops}, pages 702--703, 2020.

\bibitem{motivation}
Aakash Deep, Aashutosh Litoriya, Akshay Ingole, Vaibhav Asare, Shubham~M Bhole, and Shantanu Pathak.
\newblock \href{https://ieeexplore.ieee.org/abstract/document/9908995/}{Realtime sign language detection and recognition}.
\newblock In {\em 2022 2nd Asian Conference on Innovation in Technology (ASIANCON)}, pages 1--4. IEEE, 2022.

\bibitem{desai2024slbias}
Aashaka Desai, Maartje De~Meulder, Julie~A Hochgesang, Annemarie Kocab, and Alex~X Lu.
\newblock \href{https://arxiv.org/abs/2403.02563}{Systemic Biases in Sign Language AI Research: A Deaf-Led Call to Reevaluate Research Agendas}.
\newblock In {\em arXiv preprint arXiv:2403.02563}, 2024.

\bibitem{gorpincenko2022extending}
Artjoms Gorpincenko and Michal Mackiewicz.
\newblock \href{https://link.springer.com/chapter/10.1007/978-3-031-25825-1_8}{Extending Temporal Data Augmentation for Video Action Recognition}.
\newblock In {\em Computer Vision -- ECCV 2022 Workshops}, pages 116--133. Springer, 2022.

\bibitem{lse}
Eva Gutierrez-Sigut, Brendan Costello, Cristina Baus, and Manuel Carreiras.
\newblock \href{https://link.springer.com/inproceedings/10.3758/s13428-014-0560-1}{LSE-sign: A lexical database for spanish sign language}.
\newblock In {\em Behavior Research Methods}, volume~48, pages 123--137. Springer, 2016.

\bibitem{self-mutual}
Aiming Hao, Yuecong Min, and Xilin Chen.
\newblock \href{http://openaccess.thecvf.com/content/ICCV2021/html/Hao_Self-Mutual_Distillation_Learning_for_Continuous_Sign_Language_Recognition_ICCV_2021_paper.html}{Self-mutual distillation learning for continuous sign language recognition}.
\newblock In {\em Proceedings of the IEEE/CVF international conference on computer vision}, pages 11303--11312, 2021.

\bibitem{hu2023continuous}
Lianyu Hu, Liqing Gao, Zekang Liu, and Wei Feng.
\newblock \href{http://openaccess.thecvf.com/content/CVPR2023/html/Hu_Continuous_Sign_Language_Recognition_With_Correlation_Network_CVPR_2023_paper.html}{Continuous sign language recognition with correlation network}.
\newblock In {\em Proceedings of the IEEE/CVF Conference on Computer Vision and Pattern Recognition}, pages 2529--2539, 2023.

\bibitem{Huber1964RobustEO}
Peter~J Huber.
\newblock \href{https://link.springer.com/chapter/10.1007/978-1-4612-4380-9_35}{Robust estimation of a location parameter}.
\newblock In {\em Breakthroughs in statistics: Methodology and distribution}, pages 492--518. Springer, 1992.

\bibitem{krsl}
Alfarabi Imashev, Medet Mukushev, Vadim Kimmelman, and Anara Sandygulova.
\newblock \href{https://aclanthology.org/2020.conll-1.51/}{A dataset for linguistic understanding, visual evaluation, and recognition of sign languages: The k-rsl}.
\newblock In {\em Proceedings of the 24th conference on computational natural language learning}, pages 631--640, 2020.

\bibitem{slr_widespread}
Vaishnavi Jadhav, Priyal Agarwal, Dhruvisha Mondhe, Rutuja Patil, and CS Lifna.
\newblock \href{https://www.researchgate.net/profile/Vaishnavi-Jadhav-12/publication/366327889_A_Survey_of_Sign_Language_Recognition_Systems/links/63a47330097c7832ca5906a7/A-Survey-of-Sign-Language-Recognition-Systems.pdf?origin=booktitleDetail&_tp=eyJwYWdlIjoiam91cm5hbERldGFpbCJ9}{A Survey of Sign Language Recognition Systems}.
\newblock In {\em booktitle of Innovative Image Processing}, volume~4, pages 237--246, 2022.

\bibitem{jain2023addsl}
Sanyam Jain.
\newblock \href{https://arxiv.org/abs/2305.09736}{ADDSL: hand gesture detection and sign language recognition on annotated danish sign language}.
\newblock In {\em arXiv preprint arXiv:2305.09736}, 2023.

\bibitem{msasl}
Hamid Reza~Vaezi Joze and Oscar Koller.
\newblock \href{https://arxiv.org/abs/1812.01053}{Ms-asl: A large-scale data set and benchmark for understanding american sign language}.
\newblock In {\em arXiv preprint arXiv:1812.01053}, 2018.

\bibitem{TheRuSLan}
Ildar Kagirov, Denis Ivanko, Dmitry Ryumin, Alexander Axyonov, and Alexey Karpov.
\newblock \href{https://aclanthology.org/2020.lrec-1.746/}{TheRuSLan: Database of Russian sign language}.
\newblock In {\em Proceedings of the Twelfth Language Resources and Evaluation Conference}, pages 6079--6085, 2020.

\bibitem{kan2021sign}
Jichao Kan, Kun Hu, Markus Hagenbuchner, Ah~Chung Tsoi, Mohammed Bennamoun, and Zhiyong Wang.
\newblock \href{http://openaccess.thecvf.com/content/WACV2022/html/Kan_Sign_Language_Translation_With_Hierarchical_Spatio-Temporal_Graph_Neural_Network_WACV_2022_paper.html}{Sign language translation with hierarchical spatio-temporal graph neural network}.
\newblock In {\em Proceedings of the IEEE/CVF winter conference on applications of computer vision}, pages 3367--3376, 2022.

\bibitem{slovo}
Alexander Kapitanov, Kvanchiani Karina, Alexander Nagaev, and Petrova Elizaveta.
\newblock \href{https://link.springer.com/chapter/10.1007/978-3-031-44137-0_6}{Slovo: Russian Sign Language Dataset}.
\newblock In {\em International Conference on Computer Vision Systems}, pages 63--73. Springer, 2023.

\bibitem{kinetics}
Will Kay, Joao Carreira, Karen Simonyan, Brian Zhang, Chloe Hillier, Sudheendra Vijayanarasimhan, Fabio Viola, Tim Green, Trevor Back, Paul Natsev, et~al.
\newblock \href{https://arxiv.org/abs/1705.06950}{The kinetics human action video dataset}.
\newblock In {\em arXiv preprint arXiv:1705.06950}, 2017.

\bibitem{wlasl}
Dongxu Li, Cristian Rodriguez, Xin Yu, and Hongdong Li.
\newblock \href{http://openaccess.thecvf.com/content_WACV_2020/html/Li_Word-level_Deep_Sign_Language_Recognition_from_Video_A_New_Large-scale_WACV_2020_paper.html}{Word-level deep sign language recognition from video: A new large-scale dataset and methods comparison}.
\newblock In {\em Proceedings of the IEEE/CVF winter conference on applications of computer vision}, pages 1459--1469, 2020.

\bibitem{mvitv2}
Yanghao Li, Chao-Yuan Wu, Haoqi Fan, Karttikeya Mangalam, Bo Xiong, Jitendra Malik, and Christoph Feichtenhofer.
\newblock \href{http://openaccess.thecvf.com/content/CVPR2022/html/Li_MViTv2_Improved_Multiscale_Vision_Transformers_for_Classification_and_Detection_CVPR_2022_paper.html}{Mvitv2: Improved multiscale vision transformers for classification and detection}.
\newblock In {\em Proceedings of the IEEE/CVF conference on computer vision and pattern recognition}, pages 4804--4814, 2022.

\bibitem{liebel2018auxiliary}
Lukas Liebel and Marco K{\"o}rner.
\newblock \href{https://arxiv.org/abs/1805.06334}{Auxiliary tasks in multi-task learning}.
\newblock In {\em arXiv preprint arXiv:1805.06334}, 2018.

\bibitem{uniformaugment}
Tom~Ching LingChen, Ava Khonsari, Amirreza Lashkari, Mina~Rafi Nazari, Jaspreet~Singh Sambee, and Mario~A Nascimento.
\newblock \href{https://arxiv.org/abs/2003.14348}{Uniformaugment: A search-free probabilistic data augmentation approach}.
\newblock In {\em arXiv preprint arXiv:2003.14348}, 2020.

\bibitem{Liu_2022}
Xiaolong Liu, Qimeng Wang, Yao Hu, Xu Tang, Shiwei Zhang, Song Bai, and Xiang Bai.
\newblock \href{https://ieeexplore.ieee.org/abstract/document/9854104/}{End-to-end temporal action detection with transformer}.
\newblock In {\em IEEE Transactions on Image Processing}, volume~31, pages 5427--5441. IEEE, 2022.

\bibitem{adamW}
Ilya Loshchilov and Frank Hutter.
\newblock \href{https://openreview.net/forum?id=Bkg6RiCqY7}{Decoupled Weight Decay Regularization}.
\newblock In {\em International Conference on Learning Representations}, 2019.

\bibitem{devisign}
Lu Meng and Ronghui Li.
\newblock \href{https://www.mdpi.com/1424-8220/21/4/1120}{An attention-enhanced multi-scale and dual sign language recognition network based on a graph convolution network}.
\newblock In {\em Sensors}, volume~21, page 1120, 2021.

\bibitem{min2021visual}
Yuecong Min, Aiming Hao, Xiujuan Chai, and Xilin Chen.
\newblock \href{https://openaccess.thecvf.com/content/ICCV2021/html/Min_Visual_Alignment_Constraint_for_Continuous_Sign_Language_Recognition_ICCV_2021_paper.html?ref=https://githubhelp.com}{Visual alignment constraint for continuous sign language recognition}.
\newblock In {\em Proceedings of the IEEE/CVF international conference on computer vision}, pages 11542--11551, 2021.

\bibitem{fluentsigners}
Medet Mukushev, Aidyn Ubingazhibov, Aigerim Kydyrbekova, Alfarabi Imashev, Vadim Kimmelman, and Anara Sandygulova.
\newblock \href{https://booktitles.plos.org/plosone/inproceedings?id=10.1371/booktitle.pone.0273649}{FluentSigners-50: A signer independent benchmark dataset for sign language processing}.
\newblock In {\em Plos one}, volume~17, page e0273649. Public Library of Science San Francisco, CA USA, 2022.

\bibitem{prajwal2022weaklysupervised}
KR Prajwal, Hannah Bull, Liliane Momeni, Samuel Albanie, G{\"u}l Varol, and Andrew Zisserman.
\newblock \href{https://arxiv.org/abs/2211.08954}{Weakly-supervised fingerspelling recognition in british sign language videos}.
\newblock In {\em arXiv preprint arXiv:2211.08954}, 2022.

\bibitem{components}
Alexey Prikhodko, Mikhail Grif, and Maxim Bakaev.
\newblock \href{https://link.springer.com/chapter/10.1007/978-3-030-65218-0_34}{Sign language recognition based on notations and neural networks}.
\newblock In {\em Digital Transformation and Global Society: 5th International Conference, DTGS 2020, St. Petersburg, Russia, June 17--19, 2020, Revised Selected Papers 5}, pages 463--478. Springer, 2020.

\bibitem{Pu_2020}
Junfu Pu, Wengang Zhou, Hezhen Hu, and Houqiang Li.
\newblock \href{https://dl.acm.org/doi/abs/10.1145/3394171.3413931}{Boosting continuous sign language recognition via cross modality augmentation}.
\newblock In {\em Proceedings of the 28th ACM international conference on multimedia}, pages 1497--1505, 2020.

\bibitem{lsa64}
Franco Ronchetti, Facundo~Manuel Quiroga, C{\'e}sar Estrebou, Laura Lanzarini, and Alejandro Rosete.
\newblock \href{https://arxiv.org/abs/2310.17429}{LSA64: an Argentinian sign language dataset}.
\newblock In {\em arXiv preprint arXiv:2310.17429}, 2023.

\bibitem{autsl}
Ozge~Mercanoglu Sincan and Hacer~Yalim Keles.
\newblock \href{https://ieeexplore.ieee.org/abstract/document/9210578/}{Autsl: A large scale multi-modal turkish sign language dataset and baseline methods}.
\newblock In {\em IEEE access}, volume~8, pages 181340--181355. IEEE, 2020.

\bibitem{maskfeat}
Chen Wei, Haoqi Fan, Saining Xie, Chao-Yuan Wu, Alan Yuille, and Christoph Feichtenhofer.
\newblock \href{http://openaccess.thecvf.com/content/CVPR2022/html/Wei_Masked_Feature_Prediction_for_Self-Supervised_Visual_Pre-Training_CVPR_2022_paper.html}{Masked feature prediction for self-supervised visual pre-training}.
\newblock In {\em Proceedings of the IEEE/CVF Conference on Computer Vision and Pattern Recognition}, pages 14668--14678, 2022.

\bibitem{wu2020ioubalanced}
Shengkai Wu, Jinrong Yang, Xinggang Wang, and Xiaoping Li.
\newblock \href{https://www.sciencedirect.com/science/inproceedings/pii/S0167865522000289}{Iou-balanced loss functions for single-stage object detection}.
\newblock In {\em Pattern Recognition Letters}, volume 156, pages 96--103. Elsevier, 2022.

\bibitem{XU2019351}
Wanru Xu, Zhenjiang Miao, Jian Yu, and Qiang Ji.
\newblock \href{https://www.sciencedirect.com/science/inproceedings/pii/S0925231219300189}{Action recognition and localization with spatial and temporal contexts}.
\newblock In {\em Neurocomputing}, volume 333, pages 351--363. Elsevier, 2019.

\bibitem{yun2019cutmix}
Sangdoo Yun, Dongyoon Han, Seong~Joon Oh, Sanghyuk Chun, Junsuk Choe, and Youngjoon Yoo.
\newblock \href{http://openaccess.thecvf.com/content_ICCV_2019/html/Yun_CutMix_Regularization_Strategy_to_Train_Strong_Classifiers_With_Localizable_Features_ICCV_2019_paper.html}{CutMix: Regularization Strategy to Train Strong Classifiers with Localizable Features}.
\newblock In {\em Proceedings of the IEEE/CVF International Conference on Computer Vision (ICCV)}, pages 6022--6031, 2019.

\bibitem{zhang2022actionformer}
Chen-Lin Zhang, Jianxin Wu, and Yin Li.
\newblock \href{https://link.springer.com/chapter/10.1007/978-3-031-19772-7_29}{Actionformer: Localizing moments of actions with transformers}.
\newblock In {\em European Conference on Computer Vision}, pages 492--510. Springer, 2022.

\bibitem{zhang2018mixup}
Hongyi Zhang, Moustapha Cisse, Yann~N Dauphin, and David Lopez-Paz.
\newblock \href{https://arxiv.org/abs/1710.09412}{mixup: Beyond empirical risk minimization}.
\newblock In {\em arXiv preprint arXiv:1710.09412}, 2017.

\bibitem{zhou2023glossfree}
Benjia Zhou, Zhigang Chen, Albert Clap{\'e}s, Jun Wan, Yanyan Liang, Sergio Escalera, Zhen Lei, and Du Zhang.
\newblock \href{http://openaccess.thecvf.com/content/ICCV2023/html/Zhou_Gloss-Free_Sign_Language_Translation_Improving_from_Visual-Language_Pretraining_ICCV_2023_paper.html}{Gloss-free sign language translation: Improving from visual-language pretraining}.
\newblock In {\em Proceedings of the IEEE/CVF International Conference on Computer Vision}, pages 20871--20881, 2023.

\bibitem{zhou2020spatialtemporal}
Hao Zhou, Wengang Zhou, Yun Zhou, and Houqiang Li.
\newblock \href{https://ojs.aaai.org/index.php/AAAI/inproceedings/view/7001}{Spatial-temporal multi-cue network for continuous sign language recognition}.
\newblock In {\em Proceedings of the AAAI conference on artificial intelligence}, volume~34, pages 13009--13016, 2020.

\bibitem{templocalization}
Zixin Zhu, Le Wang, Wei Tang, Ziyi Liu, Nanning Zheng, and Gang Hua.
\newblock \href{https://ojs.aaai.org/index.php/AAAI/inproceedings/view/20277}{Learning disentangled classification and localization representations for temporal action localization}.
\newblock In {\em Proceedings of the AAAI Conference on Artificial Intelligence}, volume~36, pages 3644--3652, 2022.

\bibitem{zuo2024improving}
Ronglai Zuo and Brian Mak.
\newblock \href{https://dl.acm.org/doi/abs/10.1145/3640815}{Improving continuous sign language recognition with consistency constraints and signer removal}.
\newblock In {\em ACM Transactions on Multimedia Computing, Communications and Applications}, volume~20, pages 1--25. ACM New York, NY, 2024.

\bibitem{nlaslr}
Ronglai Zuo, Fangyun Wei, and Brian Mak.
\newblock \href{http://openaccess.thecvf.com/content/CVPR2023/html/Zuo_Natural_Language-Assisted_Sign_Language_Recognition_CVPR_2023_paper.html}{Natural language-assisted sign language recognition}.
\newblock In {\em Proceedings of the IEEE/CVF Conference on Computer Vision and Pattern Recognition}, pages 14890--14900, 2023.

\end{thebibliography}
}
\clearpage
\onecolumn
\appendix
\section*{Supplementary materials}

\section{The Variable Parameters.}
\label{sec:variable}

\begin{table}[H]
\begin{center}
\centering
    \scalebox{0.8}{
    \begin{tabular}{|c|c|c|c|}
    \hline
        Aug. type & Augmentation & Magnitude & Probability \\
        \hline \hline
        \multirow{9}[14]{*}{Image Augs.} & Color Jitter & \makecell{brightness=0.1, \\ contrast=0.005, \\ saturation=0, \\ hue=0.05} & 0.5 \\
        \cline{2-4}
        & Random Noise & \makecell{amount=[0.001, 0.005], \\ type=``s\&p" (salt-and-pepper noise)} & 0.5 \\
        \cline{2-4}
        & Sharpness & \makecell{sharpness\_factor=[0.5, 2]} & 0.35 \\
        \cline{2-4}
        & Flip & 
        \makecell{blacklisted\_labels=[739, 635, 148, 636]} & 0.5 \\
        \cline{2-4}
        & Random Erasing & 
        \makecell{min\_area\_ratio=0.02, \\ max\_area\_ratio=0.33} & 0.25 \\
        \cline{2-4}
        & Image Compression & 
        \makecell{quality\_lower=80, \\ quality\_upper=100} & 0.15 \\
        \cline{2-4}
        & Downscale & 
        \makecell{scale\_min=0.4, \\ scale\_max=0.8} & 0.15 \\
        \cline{2-4}
        & CutMix & 
        \makecell{alpha=0.8} & 1 \\
        \cline{2-4}
        & MixUp & 
        \makecell{alpha=1} & 1 \\
        \hline \hline
        \multirow{5}{*}{Video Augs.} & Random Boundary Shift & shift=[-5, 5] & 1 \\
        \cline{2-4}
        & Random Drop & drop\_ratio=0.1 & 0.5 \\
        \cline{2-4}
        & Random Add & add\_ratio=0.3 & 0.25 \\
        \cline{2-4}
        & Speed Up & acceleration=2 & 0.25 \\
        \cline{2-4}
        & Slow Down & deceleration=2 & 0.25 \\
        \hline
    \end{tabular}}
    \end{center}
        \caption{The hyperparameters of image and video augmentations.}
\label{tab:img_aug_param}
\end{table}

\begin{table}[H]
\begin{center}
\centering
  \scalebox{0.8}{
  \begin{tabular}{|c|c|c|c|c|c|}
  \hline
    Model & Pretrain Task & drop path rate & LR & LR decay & batch size \\
    \hline\hline
     \multirow{2}{*}{MViTv2-S} & MaskFeat & 0.1 & 0.0096 & \cmark & 16 \\ 
    \cline{2-6}
     & \multirow{3}{*}{Classification} & 0.2 & 0.0016 & \xmark & 8 \\
    \cline{1-1} \cline{3-6}
    MViTv2-B & & 0.3 & 0.0016 & \xmark & 6 \\
    \cline{1-1} \cline{3-6}
    I3D & & --- & 0.0096 & \xmark & 16 \\
    \hline
  \end{tabular}}
  \end{center}
        \caption{The variable parameters in the experimental setups. For MViTv2-S pretrained on MaskFeat, we use layer-wise learning rate decay by 0.75 to decrease it with different speeds for different network layers. LR is abbreviated for learning rate. Other parameters remain unchanged as described in Section~\ref{sec:experiments}.}
\label{tab:exp_params}
\end{table}

\section{SlovoExt Creation Pipeline}
\label{sec:collection}

We collected the self-assembled part of the SlovoExt dataset in three main stages: (1) data mining, (2) validation, and (3) annotation. Tasks were available only to those who passed the mandatory RSL exam and could thus be considered native speakers. We utilized Yandex Toloka\footnote{https://platform.toloka.ai/} for the first stage and ABC Elementary\footnote{https://elementary.activebc.ru} for the others due to the specifics of each of the platforms and to avoid checking signers own videos. The creation pipeline of the self-assembled Russian Isolated Sign Language Dataset is briefly described below.

\begin{enumerate}
\item \textbf{Mining.} The sign basket was the same as in the Slovo dataset~\cite{slovo}: 1,000 frequently used words with a video template from the SpreadTheSign website\footnote{https://www.spreadthesign.com/ru.ru/search/} were collected. The signers were informed about further publication of submitted videos.
\item \textbf{Validation.} The recorded videos were strictly validated, excluding videos performed incorrectly and low-resolution videos. At least three crowdworkers confirmed the correctness of each video.
\item \textbf{Annotation.} Crowdworkers indicated the start and end frames of the sign on each video to train the model to locate it. Three different crowdworkers were assigned to annotate each video, and their annotations were averaged to create a high-confidence, accurate annotation.
\end{enumerate}

Then, we expand the Slovo with a self-assembled dataset and split the result SlovoExt into train (75\%) and test (25\%) subsets, ensuring minimal user overlap between them.

\section{Dataset Licenses}
\label{sec:licenses}
The WLASL~\cite{wlasl} dataset is distributed under the Computational Use of Data Agreement (C-UDA)\footnote{https://github.com/microsoft/Computational-Use-of-Data-Agreement}, restricting its usage exclusively to academic and computational purposes. Redistribution of the data is allowed if credit or attribution information is included. The AUTSL~\cite{autsl} dataset, as stated at the project site\footnote{https://cvml.ankara.edu.tr/datasets/}, was developed and made available exclusively for academic and research purposes. The Slovo~\cite{slovo} dataset is licensed under a variant of Creative Commons Attribution-ShareAlike 4.0 International License\footnote{https://creativecommons.org/licenses/by-sa/4.0/deed.en}. Anyone can share and adapt this data for any purpose, even commercially.

\end{document}